\documentclass[runningheads]{llncs}

\usepackage{eccv}
\usepackage{gensymb}
\usepackage{multirow}
\usepackage{wrapfig}
\usepackage{hyperref}
\usepackage[hyphenbreaks]{breakurl}

\usepackage{eccvabbrv}

\usepackage{graphicx}
\usepackage{booktabs}
\usepackage{wrapfig}
\usepackage{tikz}
\usetikzlibrary{positioning, arrows.meta, calc, decorations.pathreplacing}

\usepackage[accsupp]{axessibility}  

\usepackage{hyperref}
\usepackage[misc]{ifsym}

\usepackage{orcidlink}

\makeatletter
\DeclareRobustCommand\onedot{\futurelet\@let@token\@onedot}
\def\@onedot{\ifx\@let@token.\else.\null\fi\xspace}

\def\etal{et~al\onedot}

\makeatother

\newcommand{\boldparagraph}[1]{\vspace{0.2cm}\noindent{\bf #1:} }

\definecolor{darkgreen}{rgb}{0,0.7,0}

\usepackage{pifont}

\newcommand{\gain}[1]{{\color{darkgreen}\tiny(#1)}}
\newcommand{\loss}[1]{{\color{red}\tiny(#1)}}

\newcommand{\samethanks}[1][\value{footnote}]{\footnotemark[#1]}

\begin{document}

\title{UniFusion: Sparse-View 4D Reconstruction via Unified Spatio-temporal Depth Alignment}
\renewcommand{\thefootnote}{\fnsymbol{footnote}}
\titlerunning{UniFusion}
\newcommand{\corrauth}{\textsuperscript{\dagger}}
\author{
Yongzhe Lyu\thanks{These authors contributed equally.~~~\Letter~Corresponding authors.}\orcidlink{0009-0000-8420-9538}
\and
Shaofei Wang\samethanks\orcidlink{0000-0002-2865-698X}
\and
Yixin Chen$^{\text{\Letter}}$\orcidlink{0000-0002-8176-0241}
\and
Siyuan Huang$^{\text{\Letter}}$\orcidlink{0000-0003-1524-7148}
}

\authorrunning{Lyu \etal}

%
%
\institute{State Key Laboratory of General Artificial Intelligence, BIGAI, Beijing, China 
}

\maketitle


\begin{abstract}
In this paper, we address the challenging problem of 4D reconstruction from sparse-view videos. This setup usually relies on monocular depth estimation to provide priors for the reconstruction model. A key challenge arises from limited cross-view overlap and temporal variation, making monocular depth predictions inconsistent across views and time.
Existing methods align spatial and temporal dimensions in separate stages, requiring foreground segmentation masks while failing to leverage temporal cues for cross-view alignment.
Contrary to these methods, we propose a unified spatial-temporal depth alignment framework that jointly resolves cross-view and cross-time inconsistencies without distinguishing foreground/background.
Our method represents depth maps across views and time as a set of spatio-temporal neural fields. This representation not only yields fast convergence, but also captures spatio-temporal correlation among depth maps implicitly, without dependence on external segmentation/tracking models.
We also propose a multi-view depth-order loss while leveraging the classic scale-and-shift-invariant loss to further improve the final depth quality.
The aligned depths initialize and supervise Gaussian splatting models for 4D reconstruction.
Experiments on Ego-Exo4D and EgoHuman demonstrate that our improved depth alignment substantially benefits dynamic Gaussian-splatting-based reconstruction methods for novel-time/view synthesis and geometry accuracy/consistency.
\end{abstract}


%
\section{Introduction}
\label{sec:intro}
\begin{figure}[t]
    \centering
    \includegraphics[width=\linewidth]{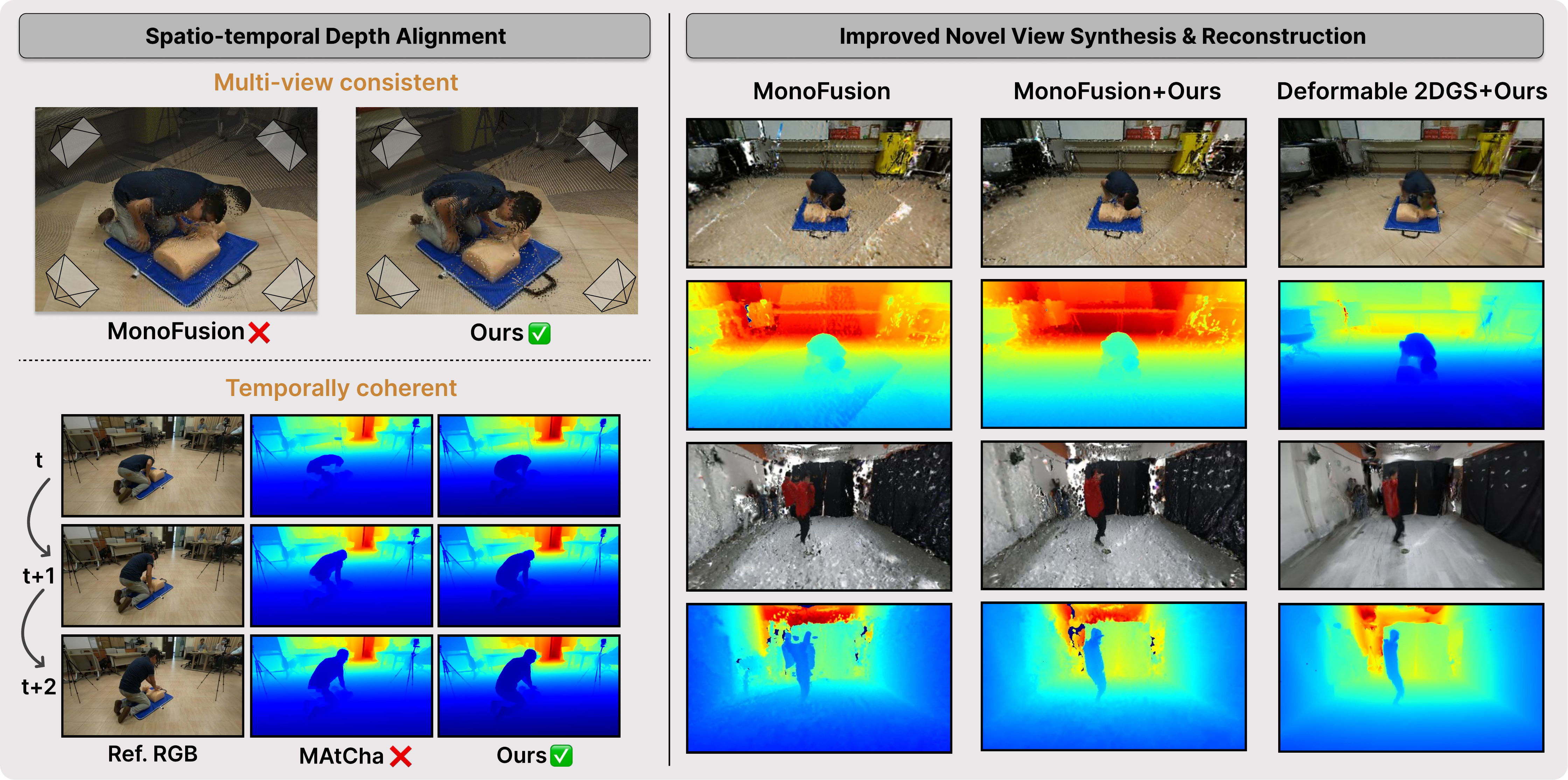}
    \caption{\textbf{Sparse-view 4D Reconstruction via spatially and temporally aligned monocular depths.} \textit{Left:} given 4-view video streams with minimal overlap, we leverage a unified spatio-temporal depth alignment framework to produce multi-view consistent (top) and temporally coherent (bottom) depth maps. \textit{Right}: the aligned depth maps serve as both initialization and supervision for 4D reconstruction, substantially improving the quality over state-of-the-art methods, achieving reasonable novel view synthesis even when shifted $\sim 45^\circ$ from the training view.}
    \label{fig:teaser}
\end{figure}
Reconstructing dynamic 3D scenes from multi-view videos is of broad interest for augmented/virtual reality, robotics, and content creation.
Dense multi-view capture systems~\cite{Collet2015SIGGRAPH,Guo2019SIGGRAPH,Edoardo2022SIGGRAPH,Icsik2023TOG} employ hundreds of synchronized cameras, achieving impressive reconstruction quality, but require expensive, lab-confined infrastructure.
This has motivated a strong practical push toward in-the-wild capture with only a few portable cameras, as exemplified by the Ego-Exo4D dataset~\cite{Grauman2024CVPR}, which uses three to four cameras to record diverse human activities.
The central tension is clear: practitioners want the reconstruction quality of dense-camera studios from the convenience of sparse-camera setups.
Crucially, this sparse-view setting differs from typical ``sparse-view'' benchmarks where a set of forward-facing cameras shares a large overlap; in our setting, cameras are separated by roughly $90^\circ$, providing complete scene coverage but limited cross-view correspondences.

Recent monocular depth and geometry estimators~\cite{Ke2024CVPR,Hu2024PAMI,Wang2025CVPRc,Wang2025NeurIPS,Lin2026ICLR} predict remarkably detailed per-view geometry from single images.
These foundation models provide strong geometric priors for sparse-view reconstruction.
However, monocular depth predictions carry unknown per-view scale/shift and are inconsistent across both views and time.
Naively merging them produces chaotic, duplicated geometry with unresolvable artifacts.
The central challenge is therefore \emph{alignment}: bringing independently predicted monocular depth maps into a consistent spatio-temporal domain.
This is fundamentally harder than the static case, because genuine scene motion is entangled with depth prediction errors, making inconsistencies across views and time difficult to disentangle.

MonoFusion~\cite{Wang2025ICCVa} addresses sparse-view dynamic reconstruction by carefully fusing monocular predictions with sparse SfM points~\cite{Wang2024CVPR}, achieving promising results.
However, it handles spatial alignment
and temporal alignment
as separate stages.
This separation introduces three drawbacks.
First, it uses only background for spatial alignment; this requires foreground masks~\cite{Kirillov2023ICCV,Ravi2024ICLR,Carion2026ICLR} to distinguish static from dynamic regions, adding a fragile dependency on an external segmentation model.
Second, temporal alignment heavily depends on tracking models~\cite{Doersch2022NeurIPS,Doersch2023ICCV,Vecerik2023ICRA,Doersch2024ACCV,Koppula2024NeurIPS,Zholus2025ICCV}, which not only result in prolonged preprocessing time but also introduce additional error accumulation.
Third, the independent stages fail to leverage temporal cues for improving cross-view alignment and vice versa.
Meanwhile, MAtCha~\cite{Guedon2025CVPR} has also demonstrated effective monocular depth map alignment (which they term as ``chart alignment'') for static sparse-view reconstruction, but does not address temporal variation.

In this work, we propose \emph{unified spatio-temporal depth alignment}: a single framework that jointly optimizes monocular depth alignment across all views and time steps.
Our key idea is to represent depth maps across views and time as a set of spatio-temporal neural fields, where all temporal frames within each view share the same spatial representation augmented with a temporal encoding for time-dependent variation.
This shared representation provides compactness and smoothness priors that yield fast convergence, while naturally capturing spatio-temporal correlation among depth maps without explicit motion modeling.
Crucially, this formulation treats all regions uniformly: the representation learns what varies temporally without being told, eliminating the need for foreground segmentation and pixel tracking, as well as making the method more robust and simpler to deploy.
We further propose a multi-view depth-order loss and leverage the classic scale-and-shift-invariant loss to improve the final depth quality.

We instantiate 4D reconstruction with our spatio-temporally aligned depths and demonstrate consistent improvements over different Gaussian splatting variants. 
Notably, we demonstrate that, when initialized and supervised properly, straightforward Gaussian splatting frameworks~\cite{Wu2024CVPR,Huang2024SIGGRAPH} substantially outperform existing complex multi-stage processing pipelines~\cite{Wang2025ICCVa}.

In summary, our contributions are:
\begin{itemize}
  \item \textbf{Unified alignment framework.}
    We propose to use spatio-temporal neural fields for depth alignment, jointly optimizing monocular depth alignment across all views and time steps in a single model, replacing the separate spatial and temporal alignment stages of prior work.
  \item \textbf{Minimal external dependencies.}
    Our method handles static and dynamic regions uniformly without distinguishing foreground and background, while capturing motion implicitly without relying on external tracking models. These are natural consequences of the unified formulation rather than adding architectural complexity.
  \item \textbf{Empirical validation.}
    The aligned depths serve as initialization and supervision for 4D Gaussian splatting, providing consistent improvement for the novel-view/time synthesis (PSNR $+4 \sim 6$dB) and geometric quality (absolute relative error $-14\% \sim 40\%$) on Ego-Exo4D and EgoHuman.
\end{itemize}
Code is available at \href{https://yongzhelyu.github.io/UniFusion/}{https://yongzhelyu.github.io/UniFusion/}.


\section{Related Work}
\label{sec:related}

\boldparagraph{Monocular depth estimation}
Monocular depth estimation has advanced rapidly, progressing from early CNN-based approaches~\cite{Eigen2014NeurIPS} to modern foundation models with strong zero-shot generalization.
MiDaS~\cite{Ranftl2022PAMI} and Omnidata~\cite{Eftekhar2021ICCV} established the paradigm of training on diverse, large-scale datasets---the former by mixing existing depth datasets for zero-shot cross-dataset transfer, the latter by generating consistent annotations from 3D scans at scale.
Subsequent discriminative methods further improved accuracy: ZoeDepth~\cite{Bhat2023ARXIV} bridges relative and metric depth through a two-stage architecture, and Metric3D~v2~\cite{Hu2024PAMI} incorporates camera intrinsics to resolve metric ambiguity.
In parallel, diffusion-based approaches such as Marigold~\cite{Ke2024CVPR} fine-tune latent diffusion models on synthetic depth data, achieving strong generalization without real-world depth labels.
The Depth Anything series~\cite{Yang2024CVPRb,Yang2024NeurIPS} scales monocular depth training, first through large-scale self-training on unlabeled images, then by distilling a synthetic-data teacher via pseudo-labeled real images. MoGe~\cite{Wang2025CVPRc} takes a complementary approach, predicting affine-invariant 3D point maps rather than depth to eliminate focal-distance ambiguity, with its successor~\cite{Wang2025NeurIPS} further recovering metric scale depth.
Despite these progresses, a fundamental limitation persists: each image is processed independently, yielding per-view predictions with unknown scale and shift.
While video depth methods such as DepthCrafter~\cite{Hu2024DepthCrafter} and Geo4D~\cite{Jiang2025ICCV} repurpose video diffusion models for temporally consistent depth sequences, they do not address multi-view geometric consistency.
Aligning these independently predicted depths across views and over time remains the central challenge that motivates our work.

\boldparagraph{Multi-view geometry and depth alignment}
Classical structure from motion (SfM)~\cite{Hartley2000Book,Oliensis2000CVIU,Ozyesil2017Survey}, exemplified by COLMAP~\cite{Schonberger2016CVPR}, produces sparse but globally consistent 3D reconstructions.
Recent learning-based methods relax these requirements: DUSt3R~\cite{Wang2024CVPR} and MASt3R~\cite{Leroy2024MASt3R} predict dense pointmaps from image pairs without explicit camera calibration, VGGT~\cite{Wang2025CVPRa} predicts cameras, point maps, and depth from unposed images in a single forward pass. Although these methods produce multi-view consistent point maps, the point maps are still relatively coarse and lack fine details, compared to dedicated monocular depth prediction models.
MAtCha~\cite{Guedon2025CVPR} introduces chart alignment---fitting per-view neural deformation fields to jointly align monocular depths~\cite{Yang2024CVPRb} and point maps~\cite{Leroy2024MASt3R} toward multi-view consistency, achieving high-quality static geometry from sparse views.
However, it does not address dynamic reconstruction.
No existing method jointly aligns monocular depths across both views and time; we extend chart alignment to the spatio-temporal domain to fill this gap.

\boldparagraph{Dynamic scene reconstruction}
Neural radiance fields~\cite{Mildenhall2020ECCV} enabled photorealistic novel-view synthesis, and subsequent extensions model dynamic scenes through learned deformation fields or direct temporal conditioning~\cite{Li2022CVPR,Attal2023CVPR,Icsik2023TOG,Kim2024AAAI,Peng2023CVPR,Song2023TVCG,Wang2023ICCV,Wang2023SIGGRAPHAsia}.
D-NeRF~\cite{Pumarola2021CVPR}, Nerfies~\cite{Park2021ICCV}, and HyperNeRF~\cite{Park2021TOG} pioneered this direction but typically require dense multi-view input or moving cameras and are slow to train.
Efficient spatio-temporal representations such as HexPlane~\cite{Cao2023CVPR}, K-Planes~\cite{FridovichKeil2023CVPR}, and Tensor4D~\cite{Shao2023CVPR} factorize 4D volumes into planar or voxel components for fast training, while ResFields~\cite{Mihajlovic2024ICLR} adds temporal conditioning through weight residuals.
The advent of 3D Gaussian splatting~\cite{Kerbl2023TOG} has brought real-time rendering to dynamic scenes.
Deformation-based methods~\cite{Yang2024CVPRa,Bae2024ECCV,Liu2025ICLR,Guo2025TCSVT,Lu2024CVPR,Shaw2024ECCV,Zhu2024NeurIPS,Labe2024ECCV,Liang2025WACV,Xu2024NeurIPS,Kim2024NeurIPS,Xu2024CVPR,Wang2025ICCVb} employ explicit motion models to adjust Gaussian parameters over time, while 4D-primitive approaches~\cite{Yang2024ICLR,Wu2024CVPR,Li2024CVPR,Lee2024NeurIPS,Luiten20243DV,Duan2024SIGGRAPH,Wang2025CVPRb,Xu2024TOG,Gao2025ARXIV} instead directly model dynamics with temporally extended Gaussian primitives.
Sparse-view dynamic reconstruction remains severely under-explored: SplatFields~\cite{Mihajlovic2024ECCV} addresses sparse-view synthesis but focuses on foreground objects without background modeling.
MonoFusion~\cite{Wang2025ICCVa} achieves full dynamic scene reconstruction by fusing monocular estimates, but relies on separate spatial and temporal alignment stages with additional foundational model dependencies~\cite{Ravi2024ICLR,Koppula2024NeurIPS,Oquab2024TMLR} aside from monocular depth and SfM points.
We demonstrate that unified spatio-temporal depth alignment, combined with a straightforward deformable Gaussian splatting pipeline, outperforms more complex multi-stage approaches, underscoring that effectively leveraging depth priors matters more than architectural complexity.


\section{Method}
\label{sec:method}

\begin{figure}[t]
  \centering
  \resizebox{0.95\linewidth}{!}{\input{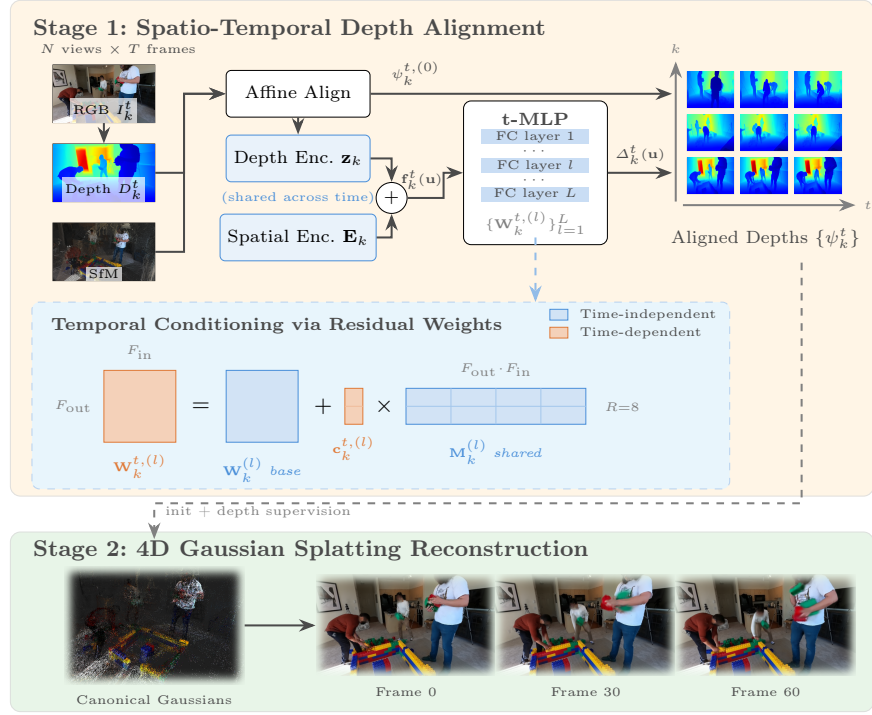}}
  \caption{\textbf{Overview of our method.}
    Given monocular depths (estimated from RGB images) and sparse SfM points from sparse synchronized cameras (\emph{left}),
    we align them jointly across views and time via per-view spatio-temporal
    neural fields (\emph{Stage~1}).
    Each view maintains spatial and depth encodings shared across time, while a per-view deformation MLP predicts depth
    offsets along ray directions.
    Temporal variation is captured through low-rank residual weight conditioning
    (\emph{inset}): each layer's weights decompose into a shared base plus a
    compact residual parameterized by per-frame temporal coefficients.
    The aligned depths initialize and supervise 4D Gaussian splatting
    for novel-view/time synthesis (\emph{Stage~2}).}
  \label{fig:pipeline}
\end{figure}

Given $N$ synchronized cameras with either known or estimated~\cite{Wang2024CVPR,Leroy2024MASt3R,Wang2025CVPRa} intrinsics and extrinsics observing a dynamic scene over $T$ time steps, our goal is to reconstruct the 4D scene for novel-view/time synthesis.

Specifically, we leverage a unified spatio-temporal depth alignment framework to provide strong geometric initialization and supervision for 4D reconstruction. It jointly corrects all monocular depth maps across views and time through a neural-field-based deformation model (\cref{sec:representation}), optimized with multi-view geometric losses (\cref{sec:optimization}). The aligned depths then initialize and supervise subsequent Gaussian splatting reconstruction approaches (\cref{sec:downstream}).
An overview is shown in \cref{fig:pipeline}.


\subsection{Spatio-Temporal Depth Alignment}
\label{sec:representation}

Our key idea is to represent all depth maps across views and time as a set of spatio-temporal neural fields.
Within these fields, all temporal frames of a given view share the same spatial representation, while frame-specific variation is captured by temporal conditioning in the network weights.
This formulation allows joint optimization across space and time without separating static and dynamic regions, and hence without requiring segmentation masks or point trajectories.

We build on the per-view depth parameterization of MAtCha~\cite{Guedon2025CVPR}, which was designed for \emph{static} multi-view alignment, and extend it to the spatio-temporal setting through residual temporal conditioning. 

\paragraph{Time-independent Encoding.}
For each view $k$ at time $t$, we first fit an affine transformation to coarsely align the monocular depth obtained from Depth Anything V2~\cite{Yang2024NeurIPS} to sparse SfM points from MASt3R~\cite{Leroy2024MASt3R} via weighted least squares, then backproject the aligned depth $D_k^t$ to 3D:
\begin{equation}
  \psi_k^{t,(0)}(\mathbf{u}) = \mathbf{o}_k + D_k^t(\mathbf{u}) \cdot \mathbf{r}_k(\mathbf{u}),
  \label{eq:depth_init}
\end{equation}
where $\mathbf{u} \in [0,1]^2$ are normalized pixel coordinates. $\mathbf{o}_k$ and $\mathbf{r}_k(\mathbf{u})$ are the camera center and ray directions at the $k$th view, respectively.
Following MAtCha~\cite{Guedon2025CVPR}, for each view $k$ we maintain two learnable encodings \emph{shared} across all $T$ time steps: a \emph{spatial encoding} $\mathbf{E}_k \in \mathbb{R}^{r_h \times r_w \times d}$, a sparse 2D feature grids interpolated bilinearly as $\mathbf{E}_k(\mathbf{u})$; and a \emph{depth encoding} $\mathbf{z}_k$, a 1D grid along the depth axis queried by quantile-normalized depth $\mathbf{z}_k(D_k^t(\mathbf{u}))$.
The two are combined by addition to form the time-independent encoding:
\begin{equation}
  \mathbf{f}_k^t(\mathbf{u}) = \mathbf{E}_k(\mathbf{u}) + \mathbf{z}_k(D_k^t(\mathbf{u})).
  \label{eq:features}
\end{equation}

\paragraph{Deformation MLP}
The combined encoding Eq.~\eqref{eq:features} is fed
to per-view deformation networks $\{ f_{\theta_k} \}_{k=1}^N$ to predict refined/aligned depth maps of a \emph{static} scene. With a slight abuse of notations, we temporarily remove the $t$ superscript for Eq.~\eqref{eq:features}. The deformation network $f_{\theta_k}: \mathbb{R}^d \rightarrow \mathbb{R}$ takes the input encoding $\mathbf{f}_k(\mathbf{u})$ and predicts a scalar deformation $\Delta_k(\mathbf{u})$ along the ray direction.
Concretely, each deformation network consists of three fully-connected hidden layers with 64-dimensional features and ReLU activations.
Layer $l$ of view $k$'s network computes:
\begin{equation}
  \mathbf{h}_k^{(l)} = \mathrm{ReLU}\!\bigl(\mathbf{W}_k^{(l)}\,\mathbf{h}_k^{(l-1)} + \mathbf{b}_k^{(l)}\bigr),
  \label{eq:hidden_layer}
\end{equation}
where $\mathbf{W}_k^{(l)} \in \mathbb{R}^{F_\text{out}^{(l)} \times F_\text{in}^{(l)}}$ are learnable weights and $\mathbf{h}_k^{(0)} = \mathbf{f}_k(\mathbf{u})$.
Each deformation MLP is essentially a \textit{neural field} that takes an input encoding derived from 2.5D query coordinates and predicts depth offsets $\Delta_k(\mathbf{u})$. Note that at this point, the MLP does not take temporal conditioning as inputs, which we will describe next.

\subsubsection{Temporal conditioning via residual weights.}
\label{sec:temporal}
To capture temporal variation while sharing spatial representations across time, we replace the static weights $\mathbf{W}_k^{(l)}$ in each hidden layer of the deformation MLP with time-dependent versions via per-timestep weight residuals~\cite{Mihajlovic2024ICLR}. This essentially extends the spatial neural field to a \textit{spatio-temporal} neural field, enabling joint alignment of depth maps across views and time simultaneously. The smoothness bias of neural fields also significantly reduces the temporal jittering of aligned depth maps.

Specifically, the weight matrix of the hidden layer $l$ at view $k$ decomposes into a base $\mathbf{W}_k^{(l)}$ shared across all frames plus a time-dependent residual:
\begin{equation}
  \mathbf{W}_k^{t,(l)} = \mathbf{W}_k^{(l)} + \delta \mathbf{W}_k^{t,(l)}.
  \label{eq:resfield}
\end{equation}
To keep the parameter count compact, we factor the residual via vector-matrix (VM) decomposition~\cite{Chen2022ECCV,Mihajlovic2024ICLR}:
\begin{equation}
  \delta \mathbf{W}_k^{t,(l)} = (\mathbf{c}_k^{t,(l)})^\top \mathbf{M}_k^{(l)},
  \label{eq:vm}
\end{equation}
where $\mathbf{c}_k^{t,(l)} \in \mathbb{R}^R$ are per-frame temporal coefficients and $\mathbf{M}_k^{(l)} \in \mathbb{R}^{R \times (F_\text{out}^{(l)} \cdot F_\text{in}^{(l)})}$ is a shared matrix ($R{=}8$ in practice), adding only $T \!\cdot\! R + R \!\cdot\! F_\text{out} \!\cdot\! F_\text{in}$ parameters per layer. Eq.~\eqref{eq:hidden_layer} can then be extended to handle temporal conditioning by simply querying the weight matrix with $t$:
\begin{equation}
  \mathbf{h}_k^{t,(l)} = \mathrm{ReLU}\!\bigl(\mathbf{W}_k^{t,(l)}\,\mathbf{h}_k^{t,(l-1)} + \mathbf{b}_k^{(l)}\bigr).
  \label{eq:temporal_hidden_layer}
\end{equation}
The final aligned depth is as follows:
\begin{equation}
  \psi_k^t(\mathbf{u}) = \psi_k^{t,(0)}(\mathbf{u}) + \Delta_k^t(\mathbf{u}) \cdot \mathbf{r}_k(\mathbf{u}).
  \label{eq:deformation}
\end{equation}
This design is what enables unified spatio-temporal alignment.
For static regions, the temporal residuals are free to vanish ($\delta \mathbf{W}^t \!\approx\! \mathbf{0}$), producing consistent outputs across frames automatically.
For dynamic regions, the residuals activate to produce frame-specific adjustments.
Crucially, the network learns this distinction without being told---no foreground/background segmentation is needed. It also learns motion implicitly.
This stands in contrast to prior work~\cite{Wang2025ICCVa} that requires explicit masks to extract static regions for alignment, and needs point tracking to initialize dynamic regions.

\subsection{Optimization Objectives}
\label{sec:optimization}

All views and time steps are optimized jointly.
We first describe our proposed multi-view depth-order loss and the scale-and-shift-invariant loss, then summarize the alignment losses inherited from MAtCha~\cite{Guedon2025CVPR}.

\paragraph{Multi-view depth-order loss.}
\label{par:depth_order}
When multiple views observe the same scene at a given time step, their depth maps must be mutually consistent: a 3D point visible in one view, when reprojected into another, should not lie in front of that view's recorded surface.
We enforce this geometric constraint with a cross-view depth-order loss.
For each view pair $(k, j)$ at the same time step, we unproject view $k$'s deformed depth $\psi_k^t$ to 3D and reproject it into view $j$.
Let $z_{j \leftarrow k}(\mathbf{u})$ denote the reprojected depth in view $j$, and $\hat{d}_{j}(\mathbf{u}')$ the depth read from view $j$'s depth map at the corresponding pixel $\mathbf{u}' = \pi_j(\Pi_k^{-1}(\mathbf{u}))$.
The loss penalizes violations where the reprojected point is closer than the recorded surface:
\begin{equation}
  \mathcal{L}_\text{order} = \frac{1}{|\mathcal{N}|^2\,|\mathcal{U}|}\sum_{(k,j) \in \mathcal{N}}\sum_{\mathbf{u} \in \mathcal{U}}
    \max\!\left(0,\;
      \frac{\hat{d}_{j}(\mathbf{u}') - z_{j \leftarrow k}(\mathbf{u})}
           {\hat{d}_{j}(\mathbf{u}')}
    \right),
  \label{eq:depth_order}
\end{equation}
where $\mathcal{U}$ denotes the set of all pixels, $\mathcal{N}$ denotes the set of all view pairs.
This loss provides a purely geometric, multi-view consistency signal that requires no monocular depth prior and no explicit cross-view correspondences---the depth maps themselves supply the supervisory signal through reprojection.

\paragraph{Scale-and-shift-invariant loss.}
To provide a complementary dense metric signal, we employ a scale-and-shift-invariant (SSI) loss~\cite{Ranftl2022PAMI}.
Each depth map is normalized by subtracting its median and dividing by its mean absolute deviation:
\begin{equation}
  \widetilde{d} = \frac{d - \text{median}(d)}{\text{MAD}(d)}, \qquad \text{MAD}(d) = \text{mean}\big(|d - \text{median}(d)|\big),
  \label{eq:ssi_normalize}
\end{equation}
and the loss is the L1 distance between normalized maps:
\begin{equation}
  \mathcal{L}_\text{ssi} = \big\| \widetilde{d}_\text{pred} - \widetilde{d}_\text{ref} \big\|_1.
  \label{eq:ssi_loss}
\end{equation}
While the depth-order loss enforces multi-view geometric consistency, the SSI loss captures per-view shape similarity after factoring out scale and shift, providing complementary supervision.

\paragraph{Alignment losses.}
We adopt three losses from MAtCha~\cite{Guedon2025CVPR} for aligning the deformed depths to sparse observations and preserving monocular geometry.
(i)~An \emph{SfM fitting loss} $\mathcal{L}_\text{fit}$ aligns the deformed depths to sparse 3D points from MASt3R~\cite{Leroy2024MASt3R,Wang2024CVPR}, using learnable per-pixel confidence to downweight noisy correspondences.
(ii)~\emph{Structure preservation losses} $\mathcal{L}_\text{struct}$ penalize changes in surface normals (cosine distance) and depth gradients (L1) between the initial and deformed depth maps, ensuring that global alignment corrections do not destroy fine-grained monocular geometry.
(iii)~A \emph{cross-view matching loss} $\mathcal{L}_\text{match}$ projects deformed 3D points into other views and minimizes reprojection error for multi-view consistency.

\paragraph{Regularization and total loss.}
We regularize the spatial encoding magnitudes (L2) and depth encoding smoothness (total variation).
The total objective is:
\begin{equation}
  \mathcal{L} = \mathcal{L}_\text{fit} + \lambda_\text{s} \mathcal{L}_\text{struct} + \lambda_\text{m} \mathcal{L}_\text{match} + \lambda_\text{o} \mathcal{L}_\text{order} + \lambda_\text{ssi} \mathcal{L}_\text{ssi} + \mathcal{L}_\text{reg},
  \label{eq:total_loss}
\end{equation}
where $\lambda_\text{s}$, $\lambda_\text{m}$, $\lambda_\text{o}$, and $\lambda_\text{ssi}$ are scalar weights.

\subsection{4D Gaussian Splatting Reconstruction}
\label{sec:downstream}
We reconstruct 4D scenes using Gaussian splatting initialized from the aligned depth maps. As shown in~\cref{sec:experiments}, our spatio-temporally aligned depth maps consistently improve the quality of a variety of baselines~\cite{Wu2024CVPR,Wang2025ICCVa}.

During training of Gaussian splatting, we follow depth supervision schedules that come with the method~\cite{Wang2025ICCVa} if it provides one. For the method~\cite{Wu2024CVPR} that do not come with a depth supervision schedule, we follow that of~\cite{Guedon2025CVPR} unless specified otherwise. Please find details of each Gaussian splatting approach used in the Supp. Mat.

\section{Experiments}
\label{sec:experiments}
We evaluate our approach in this section. \cref{sec:exp_setup} describes the experimental setup; \cref{sec:comparisons} compares against state-of-the-art methods; and \cref{sec:ablations} ablates our design choices.

\subsection{Experimental Setup}
\label{sec:exp_setup}
\paragraph{Datasets.}
We evaluate on two real-world, in-the-wild datasets.
\textbf{Ego-Exo4D}~\cite{Grauman2024CVPR} provides in-the-wild sparse-view captures of diverse human activities.
Following MonoFusion~\cite{Wang2025ICCVa}, we use the ExoRecon subset: six scenarios (dance, sports, bike repair, cooking, music, healthcare), each with 300 frames from four exocentric cameras separated by ${\sim}90^\circ$. Since most of these sequences contain only 4 cameras, we use every 3rd frame for training and every other two frames for validation.
\textbf{EgoHuman}~\cite{Khirodkar2023ICCV} captures multi-human activities with egocentric and exocentric views. We chose the legoassemble and fencing sequences to conduct our experiments. These two sequences contain 8 and 20 cameras, respectively, which are ideal setups for testing the model's in-the-wild novel-view synthesis capability.  
We select four exocentric cameras also with ${\sim}90^\circ$ separation and use 300 frames per sequence, while using cameras in between the training cameras for evaluation. We use ground-truth camera poses for both datasets.

\paragraph{Metrics.}
We report PSNR~($\uparrow$), SSIM~($\uparrow$), and LPIPS~($\downarrow$) to measure photometric quality of rendered images.
For Ego-Exo4D, we report both full-image metrics and dynamic-region metrics, following MonoFusion's evaluation protocol~\cite{Wang2025ICCVa}; we additionally report Absolute Relative Error (AbsRel,~$\downarrow$) for depth quality.
For EgoHuman, we evaluate novel view synthesis on held-out cameras. We also report direct evaluation metrics on depth alignment quality (reprojection consistency, flow-warped depth error, depth-order violation rate) in the Supp. Mat.

\paragraph{Baselines.}
We compare against methods reported in MonoFusion~\cite{Wang2025ICCVa}, copying their published numbers where available (marked with $\dagger$):
SOM, Dynamic 3D Gaussians (Dyn3D-GS)~\cite{Luiten20243DV}, MV-SOM, MV-SOM-SD~\cite{Wang2025ICCVb}, and MonoFusion~\cite{Wang2025ICCVa}. We note that MonoFusion's official code does not contain any evaluation script to reproduce their reported metrics. Some evaluation details, especially on the dynamic regions, are only vaguely described in their paper. We thus re-implement the evaluation script and re-evaluate MonoFusion (no $\dagger$ superscript), which may cause differences in certain metrics compared to those reported in the original MonoFusion paper. We discuss details in the Supp. Mat.

To demonstrate the effectiveness of our proposed depth alignment, we apply the aligned depth to both MonoFusion and a simple deformable 2DGS baseline.  

\subsection{Comparisons with State-of-the-Art}
\label{sec:comparisons}
\begin{figure}[tb]
    \centering
    \includegraphics[width=\linewidth]{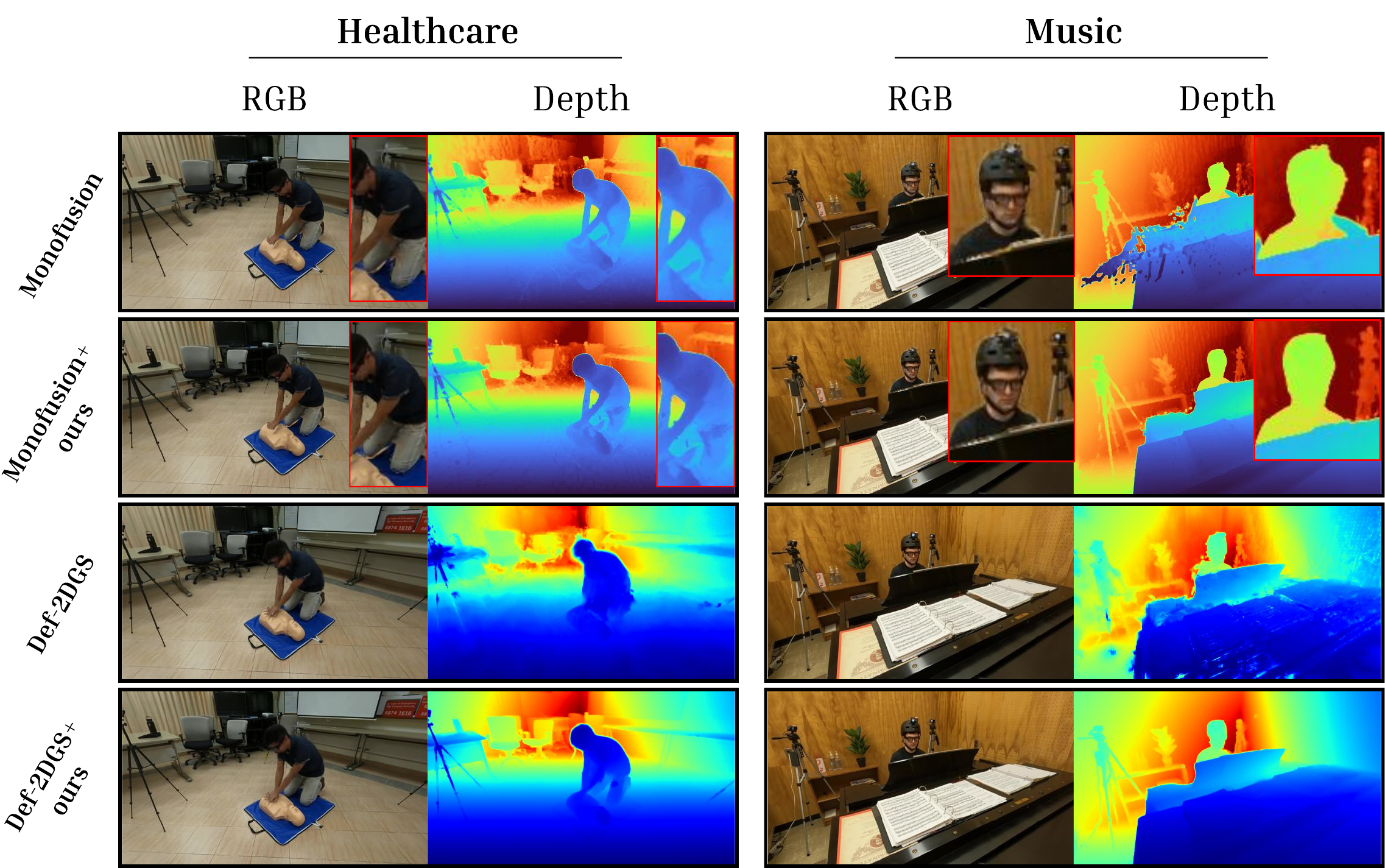}
    \caption{\textbf{Qualitative analysis of held-out frames of training view on EgoExo4D.} Our aligned depth enhances the geometry quality of both MonoFusion and deformable 2DGS, removing ghosting artifacts that are clearly visible in the held-out frames (top two rows), while substantially improving the overall depth quality of deformable 2DGS(bottom two rows).}
    \label{fig:held-out}
\end{figure}
\begin{table*}[t]
  \centering
  \caption{\textbf{Quantitative comparison on Ego-Exo4D.}
    Baseline results (marked $\dagger$) are copied from MonoFusion~\cite{Wang2025ICCVa}.
    ``X + Ours'' denotes backend X initialized and supervised with our aligned depths.}
  \label{tab:egoexo4d}
  \begin{tabular}{l|cccc|ccc}
    \toprule
    & \multicolumn{4}{c|}{Full image}
    & \multicolumn{3}{c}{Dynamic region} \\
    Method
    & PSNR~$\uparrow$ & SSIM~$\uparrow$ & LPIPS~$\downarrow$ & AbsRel~$\downarrow$
    & PSNR~$\uparrow$ & SSIM~$\uparrow$ & LPIPS~$\downarrow$ \\
    \midrule
    SOM$^\dagger$           & 14.73 & 0.535 & 0.482 & 0.843 & 15.63 & 0.559 & 0.450 \\
    Dyn3D-GS$^\dagger$      & 24.28 & 0.692 & 0.539 & 0.612 & 24.61 & 0.673 & 0.384 \\
    MV-SOM-DS$^\dagger$     & 28.37 & 0.906 & 0.079 & 0.398 & 28.23 & 0.931 & 0.063 \\
    MV-SOM$^\dagger$        & 26.91 & 0.890 & 0.138 & 0.474 & 27.31 & 0.919 & 0.078 \\
    MonoFusion$^\dagger$    & 30.43 & 0.927 & 0.061 & 0.290 & 29.71 & 0.947 & 0.017 \\
    \midrule
    MonoFusion              & 30.41 & 0.944 & 0.079 & 0.277 & 29.48 & 0.958 & 0.040 \\
    MonoFusion + Ours       & 31.63 & 0.955 & 0.064 & 0.238 & 30.44 & 0.972 & 0.032 \\[-2pt]
                            & \gain{+1.22} & \gain{+.011} & \gain{-.015} & \gain{-.039} & \gain{+.96} & \gain{+.014} & \gain{-.008} \\
    Def-3DGS                & 34.15 & 0.952 & 0.062 & 0.330 & 37.24 & 0.982 & 0.028 \\
    Def-2DGS                & 34.15 & \textbf{0.964} & 0.058 & 0.376 & 38.64 & 0.984 & 0.026 \\
    Def-2DGS + Ours         & \textbf{34.66} & 0.963 & \textbf{0.049} & \textbf{0.222} & \textbf{38.70} & \textbf{0.984} & \textbf{0.025} \\[-2pt]
                            & \gain{+.51} & \loss{-.001} & \gain{-.009} & \gain{-.154} & \gain{+.06} & & \gain{-.001} \\
    \bottomrule
  \end{tabular}
\end{table*}


\Cref{tab:egoexo4d} presents quantitative results on Ego-Exo4D for both full-image and dynamic-region metrics.
Our method consistently improves different Gaussian splatting backends.
Notably, our depth alignment substantially improves the geometry quality (AbsRel {\color{darkgreen}{$-14\%$}} for MonoFusion, {\color{darkgreen}{$-40\%$}} for deformable 2DGS). We also note that  
Def-2DGS + Ours achieves the best overall performance, significantly outperforming the state-of-the-art method (PSNR {\color{darkgreen}{$+4.25$dB}}, AbsRel {\color{darkgreen}{$-0.055$}}). The relatively smaller rendering gain over Def-2DGS is mainly because Ego-Exo4D evaluates held-out frames from training cameras, where Def-2DGS can saturate photometric metrics by overfitting on training views. 
Nevertheless, the large AbsRel reduction shows that our alignment still substantially improves geometry, and the stronger gains under held-out-camera evaluation on the EgoHuman dataset (shown later in this section) further confirm its benefit for challenging novel-view settings.

\begin{wraptable}{r}{0.6\textwidth}
  \centering
  \caption{\textbf{Quantitative comparison on EgoHuman.}
    We evaluate novel view synthesis on two sequences from the EgoHuman dataset, substantially outperforming all baselines.}
  \label{tab:egohuman}
  \begin{tabular}{l|ccc}
    \toprule
    Method
    & PSNR~$\uparrow$ & SSIM~$\uparrow$ & LPIPS~$\downarrow$ \\
    \midrule
    SOM            & 13.97 & 0.652 & 0.645 \\
    MV-SOM              & 13.36 & 0.588 & 0.564 \\
    SplatFields              & 17.74 & 0.675 & 0.571 \\
    MonoFusion              & 14.15 & 0.436 & 0.628 \\
    MonoFusion + Ours       & 20.41 & 0.612 & 0.492 \\[-2pt]
                            & \gain{+6.26} & \gain{+.176} & \gain{-.136} \\
    Def-2DGS                  & 18.88 & 0.657 & 0.431 \\
    Def-2DGS + Ours           & \textbf{22.76} & \textbf{0.856} & \textbf{0.277} \\[-2pt]
                            & \gain{+3.88} & \gain{+.199} & \gain{-.154} \\
    \bottomrule
  \end{tabular}
\end{wraptable}
\begin{figure}
    \centering
    \includegraphics[width=\linewidth]{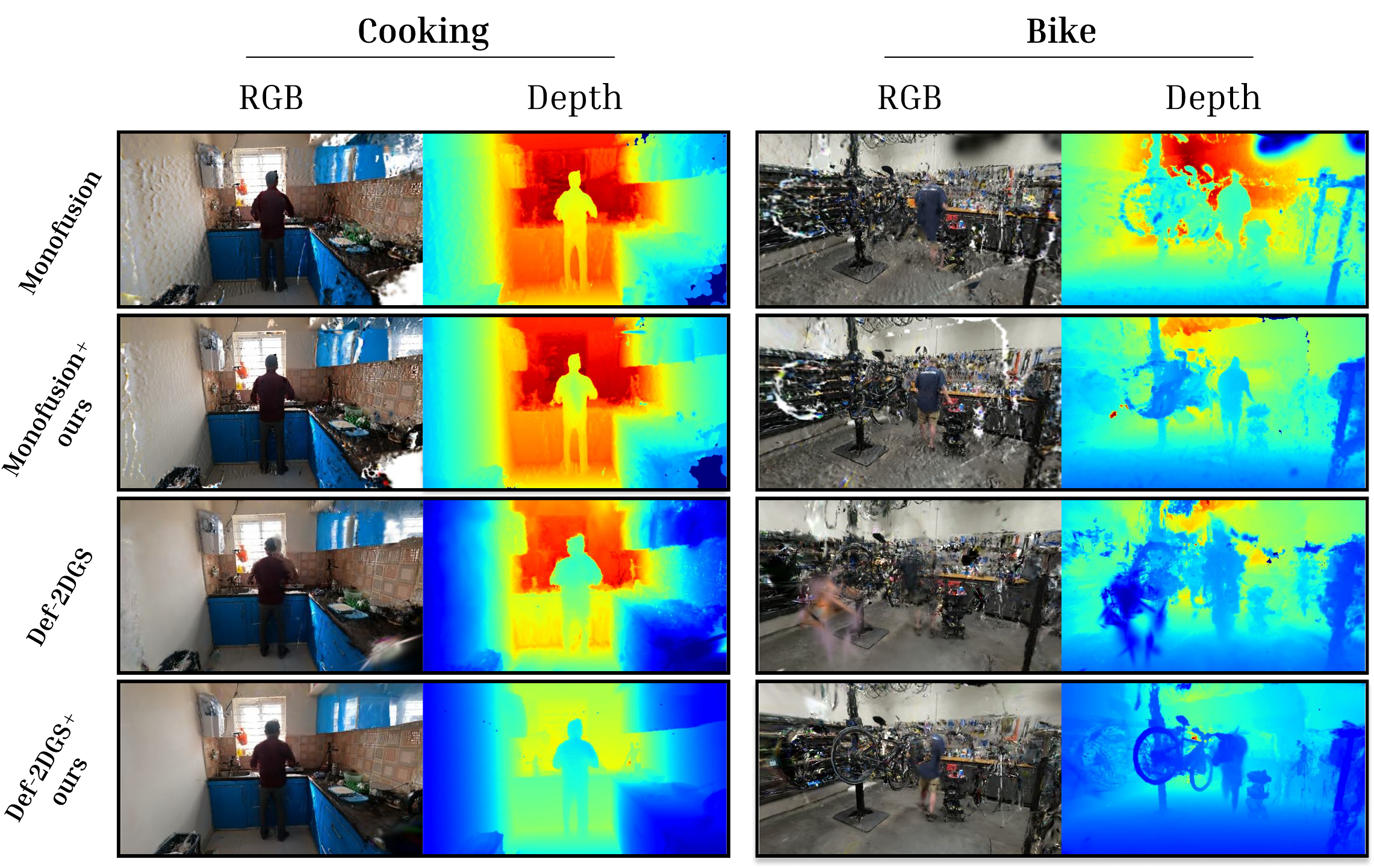}
    \caption{\textbf{Qualitative analysis of novel view synthesis on EgoExo4D.} We show extreme novel view synthesis results ($\sim 45^\circ$ from training view), further confirming that our aligned depth consistently benefits the baseline models.}
    \label{fig:light-nvs}
\end{figure}

\Cref{fig:held-out,fig:light-nvs} shows qualitative comparisons.
Our method renders sharper details and more geometrically coherent results, especially in dynamic regions where MonoFusion's separate alignment stages produce temporal artifacts. On the other hand, deformable 2DGS can overfit to training views quite well, but do not produce reasonable depth/geometry due to the lack of proper depth supervision. This is evidenced in novel view rendering, where severe artifacts manifest. In comparison, with our depth initialization and supervision, we achieve reasonable results on extreme novel views ($30^\circ \sim 45^\circ$).

\begin{figure}
    \centering
    \includegraphics[width=\linewidth]{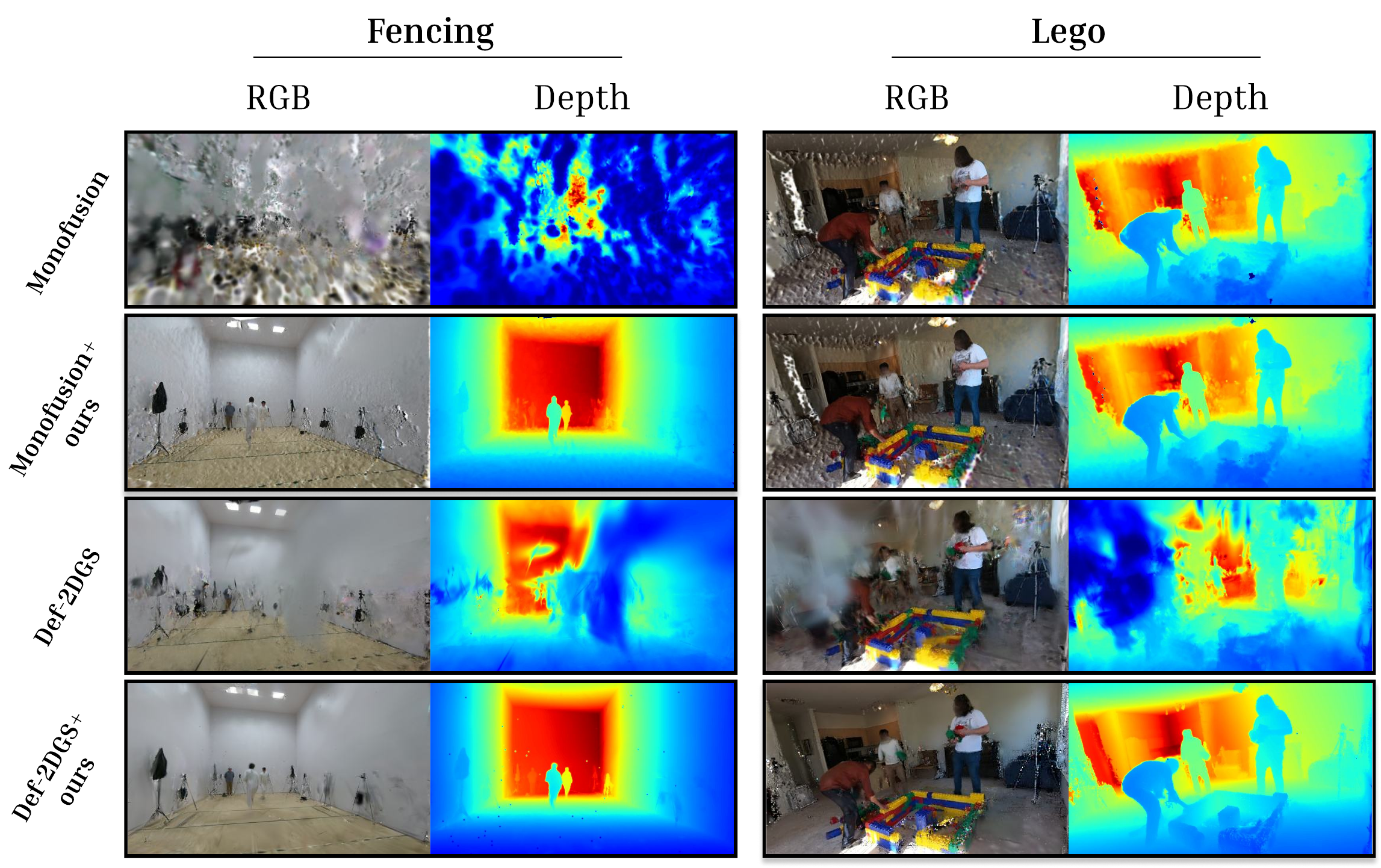}
    \caption{\textbf{Qualitative Results on the EgoHumans dataset.} On the more challenging EgoHumans dataset, our aligned depth helps MonoFusion to recover from catastrophic failure, while significantly improving the novel view rendering quality of deformable 2DGS.}
    \label{fig:egohuman}
\end{figure}

The novel view synthesis is further demonstrated quantitatively on the EgoHumans dataset~\Cref{tab:egohuman}, where significant improvements of novel view synthesis are achieved over both MonoFusion ({\color{darkgreen}{$+6.26$dB}}) and deformable 2DGS ({\color{darkgreen}{$+3.88$dB}}). Qualitative comparisons are shown in~\Cref{fig:egohuman}.

\subsection{Ablation Studies}
\label{sec:ablations}

\begin{table*}[t]
\scriptsize
  \centering
  \caption{\textbf{Progressive ablation of alignment components}
    on Ego-Exo4D on two Gaussian splatting backends.}
  \label{tab:ablation}
  \resizebox{\linewidth}{!}{%
  \begin{tabular}{l|ccc|ccc}
    \toprule
    & \multicolumn{3}{c|}{MonoFusion}
    & \multicolumn{3}{c}{Def-GS} \\
    Configuration
    & PSNR~$\uparrow$ & SSIM~$\uparrow$ & LPIPS~$\downarrow$
    & PSNR~$\uparrow$ & SSIM~$\uparrow$ & LPIPS~$\downarrow$ \\
    \midrule
    (a) Per-frame MAtCha                  & 30.14 & 0.943 & 0.070 & 32.60 & 0.956 & 0.071 \\
    (b) + time embedding             & 28.53 & 0.935 & 0.088 & 32.60 & 0.947 & 0.071 \\
    (c) + residual weights         & 30.90 & 0.948 & 0.067 & 33.61 & 0.960 & 0.055 \\
    (d) + $\mathcal{L}_\text{order}$      & 31.52 & 0.953 & 0.065 & 33.90 & 0.961 & 0.055 \\
    (e) + $\mathcal{L}_\text{ssi}$ (full) & 31.63 & 0.955 & 0.064 & 34.66 & 0.963 & 0.049 \\
    \bottomrule
  \end{tabular}}
\end{table*}

\Cref{tab:ablation} progressively ablates our design choices on Ego-Exo4D across different backends.

\paragraph{(a) Per-frame alignment} applies MAtCha's depth/chart alignment~\cite{Guedon2025CVPR} independently per frame, providing cross-view consistency without temporal modeling. 

\paragraph{(b) Time embedding} ablates a naive, temporal-embedding-conditioned MLP extension to (a). This naive implementation results in severe quality degradation.

\paragraph{(c) Temporal residual weights} incorporates our proposed residual temporal weight (~\cref{sec:temporal}). This gives the single largest improvement. We also note that the chart/depth alignment step of MAtCha over 100 training frames from 4 views takes about half a day, due to the time-consuming per-frame optimization. In contrast, our spatial-temporal alignment takes only one hour to align the same amount of data, while also achieving improved quality.

\paragraph{(d) Depth order loss} adds $\mathcal{L}_\text{order}$ (Eq.~\eqref{eq:depth_order}), which regularizes relative depth orderings and yields further improvement.

\paragraph{(e) Scale-and-shift invariant loss} completes the model with $\mathcal{L}_\text{ssi}$ (Eq.~\eqref{eq:ssi_loss}), bringing additional gains.

The trends are consistent for both MonoFusion and deformable 2DGS, confirming that our design choices are complementary to the downstream 4D reconstruction method. Additional ablations, including those on 1) the EgoHuman dataset, 2) residual rank and depth order penalty designs, and 3) different camera baselines (wider/narrower), can be found in the Supp. Mat.


\section{Conclusion}
\label{sec:conclusion}
\begin{figure}[tb]
    \centering
    \includegraphics[width=0.9\linewidth]{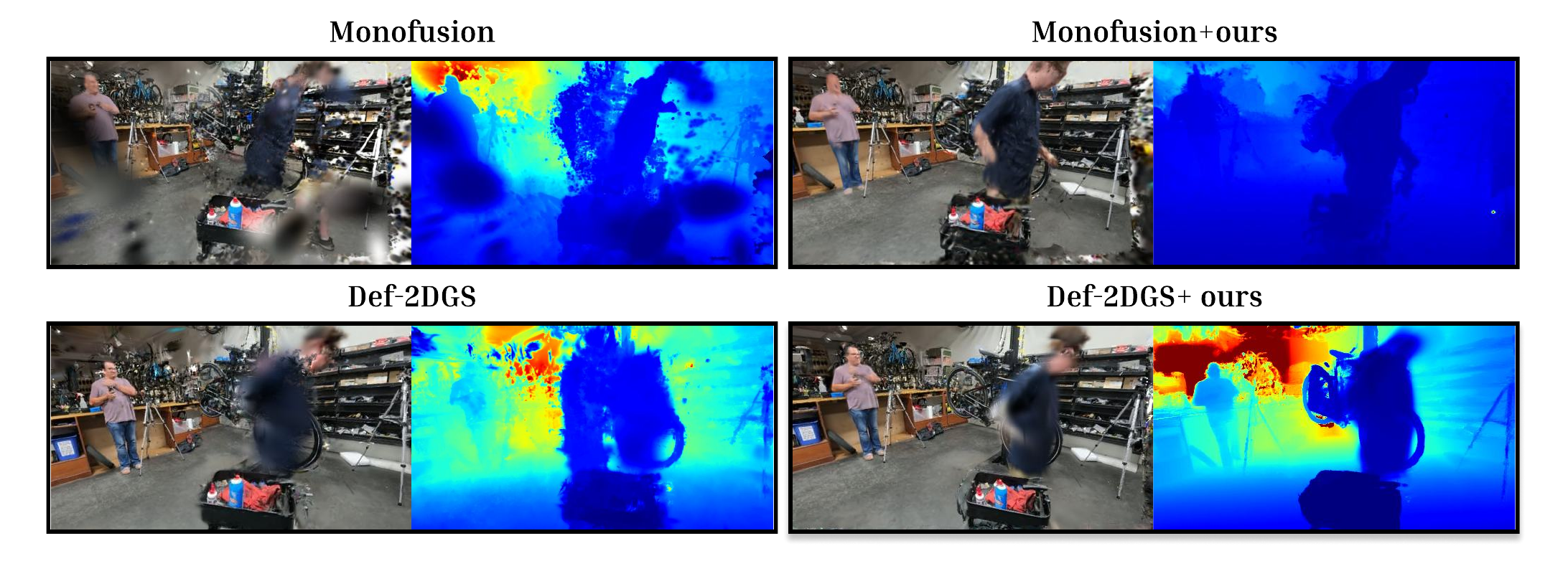}
    \caption{\textbf{Limitations.} While our improved depth consistently enhanced the geometry quality of the different baselines, we note that the naive deformable 2DGS can sometimes become the bottleneck and fail to capture the fast/complex motion. In such a case, MonoFusion, when combined with proper depth initialization/supervision from our method, can produce more reasonable foreground geometry and rendering.}
    \label{fig:limitation}
\end{figure}
We presented UniFusion, a unified spatio-temporal depth alignment framework for sparse-view 4D reconstruction. By representing per-view depth corrections as spatio-temporal neural fields with temporal residual weight conditioning, our method jointly resolves cross-view and cross-time depth inconsistencies in a single optimization, eliminating the need for separate alignment stages, foreground segmentation masks, or pixel tracking models. The aligned depths serve as both initialization and supervision for downstream Gaussian splatting reconstruction.

Experiments on Ego-Exo4D and EgoHuman demonstrate that our alignment consistently improves multiple Gaussian splatting backends, reducing geometric error (AbsRel) by 14--40\% and improving novel-view synthesis by 4--6\,dB PSNR, while being about 10$\times$ faster than naive per-frame spatial alignment. These results highlight that accurately aligned geometric priors are more impactful than downstream architectural complexity for sparse-view 4D reconstruction.

\paragraph{Limitations.}
Our framework relies on monocular depth foundation models and sparse SfM points for initialization; reconstruction quality is therefore bounded by the accuracy of these priors. We also note that the basic deformable 2DGS can fail due to fast and complex motion; in such a case, point tracking information could still help (\Cref{fig:limitation}). Promising future directions include further accelerating the alignment process for scalable 4D reconstruction, as well as integrating our alignment module with feed-forward multi-view foundational methods for real-time applications.

\newpage
\appendix

\section{Implementation Details}
\subsection{Gaussian Splatting Backends}
For 4D reconstruction, we employ two distinct backends: \textbf{Deformable 2DGS} and \textbf{MonoFusion}~\cite{Wang2025ICCVa}. 

\paragraph{Deformable 2DGS.}
Our deformable 2DGS implementation is derived from deformable 3DGS~\cite{Wu2024CVPR}, which consists of a canonical scene representation and a deformation network. In the canonical space, Gaussian primitives are used as scene representation. Each primitive $g_k$ is parameterized by its spatial position (center) $\mathbf{t}_k \in \mathbb{R}^3$, opacity $o_k \in [0, 1]$, and covariance matrix $\Sigma_k$. To ensure $\Sigma_k$ remains positive semi-definite during optimization, it is factorized into a rotation matrix $\mathbf{R}_k \in SO(3)$, typically represented via a quaternion $\mathbf{q}_k$, and a per-axis scaling matrix $\mathbf{S}_k = \text{diag}(\mathbf{s}_k)$, such that:
\begin{equation}
    \Sigma_k = \mathbf{R}_k \mathbf{S}_k \mathbf{S}_k^\top \mathbf{R}_k^\top.
\end{equation}
The spatial influence of the $k$-th Gaussian at a 3D location $\mathbf{x}$ is defined by an unnormalized Gaussian kernel $\mathcal{G}_k$:
\begin{equation}
    \mathcal{G}_k (\mathbf{x}) = \exp \left( -\frac{1}{2} (\mathbf{x} - \mathbf{t}_k)^\top \Sigma_k^{-1} (\mathbf{x} - \mathbf{t}_k) \right).
\end{equation}
When rendering, the Gaussians are projected onto the 2D image plane using EWA splatting~\cite{zwicker2002ewa}. The final color $\mathbf{C}$ at pixel $(u, v)$ is accumulated via $\alpha$-blending of $N$ depth-sorted Gaussians:
\begin{equation}
    \mathbf{C}(u, v) = \sum_{k=1}^{N} \mathbf{c}_k \alpha_k \prod_{j=1}^{k-1} (1 - \alpha_j),
\end{equation}
where $\alpha_k = o_k \mathcal{P}(\mathcal{G}_k, u, v)$ denotes the effective opacity modulated by the projected Gaussian $\mathcal{P}$, and $\mathbf{c}_k$ is the color of the Gaussian under the current viewing direction.
To enable geometric supervision (depth, normal, curvature), we use 2DGS~\cite{Huang2024SIGGRAPH} instead of 3DGS to represent the canonical Gaussians.

To deform the canonical Gaussians, ~\cite{Wu2024CVPR} uses a deformation network which consists of multi-resolution 4D K-Planes~\cite{FridovichKeil2023CVPR} and a linear projection layer to encode spatio-temporal features:
\begin{equation}
    \begin{aligned}
    \mathbf{f}_d &= \phi_d ( \bigcup_l \prod \text{interp}(R_l(i, j))), \\
    \text{s.t.} \quad & (i, j) \in \{ (x,y),(x,z),(y,z),(x,t),(y,t),(z,t)\} \\
    & l \in \{ 1, \dots, L \}
    \label{eqn:4d_decomposition}
    \end{aligned}
\end{equation}
where $\text{interp} (\cdot)$ denotes bilinear interpolation, $\bigcup$ denotes concatenation of features. $R_l(i,j)$ denotes the 4D K-Planes of resolution l, $\phi_d$ denotes the linear layer, $(x,y,z)$ denotes the position of the queried Gaussian, and $t$ denotes the target timestep. Separate decoder MLPs $\{\phi_t,\phi_q, \phi_s\}$ are then used to compute deformation deltas of position $\Delta \mathbf{t} = \phi_t(\mathbf{f}_d)$, rotation $\Delta \mathbf{q} = \phi_q(\mathbf{f}_d)$ and scaling $\Delta \mathbf{s} = \phi_s(\mathbf{f}_d)$. The deformed Gaussian $g^{'}_k$ can be obtained by $g^{'}_k=\{ \mathbf{t}_k + \Delta \mathbf{t}, \mathbf{s}_k + \Delta \mathbf{s}, \mathbf{q}_k + \Delta \mathbf{q}, o_k, \mathbf{c}_k \}$.

\paragraph{MonoFusion.}
Similar to deformable 3DGS/2DGS, MonoFusion also models dynamic scenes by decoupling the representation into a canonical 3D Gaussian field and a deformation model.

Specifically, in the canonical space at time $t_c$, similar to the standard 3DGS formulation, each primitive $g_k$ is parameterized by its spatial position $\mathbf{t}_k \in \mathbb{R}^3$, rotation $\mathbf{R}_k \in SO(3)$, per-axis scaling $\mathbf{s}_k$, opacity $o_k$, and color $\mathbf{c}_k$. Furthermore, to semantically group moving parts, each Gaussian is augmented with a semantic feature vector $\mathbf{f}_k \in \mathbb{R}^N$ ($N=32$) from 2D foundation models~\cite{Oquab2024TMLR}.

Unlike continuous MLP-based deformation networks, MonoFusion models dynamic movements through a linear combination of discrete, rigid motion bases 
. Let $\{T_{t_c \rightarrow t}^{(b)}\}_{b=1}^B$ denote a set of $B$ learnable basis trajectories in $SE(3)$ 
. The rigid transformation $T_{t_c \rightarrow t}^{(k)} = [\mathbf{R}_{t_c \rightarrow t}^{(k)} | \mathbf{t}_{t_c \rightarrow t}^{(k)}]$ for the $k$-th Gaussian from canonical space to time $t$ is formulated as a weighted sum of these bases:
\begin{equation}
    T_{t_c \rightarrow t}^{(k)} = \sum_{b=1}^{B} w^{(k,b)} T_{t_c \rightarrow t}^{(b)},
    \label{eqn:motion_bases}
\end{equation}
where the blending weights $w^{(k,b)}$ are fixed per-point coefficients initialized via K-means clustering on the semantic features $\mathbf{f}_k$ 
. The deformed position $\mathbf{t}_k(t)$ and rotation $\mathbf{R}_k(t)$ of the Gaussian at time $t$ are then obtained by applying the blended rigid transformation:
\begin{equation}
    \begin{aligned}
        \mathbf{t}_k(t) &= \mathbf{R}_{t_c \rightarrow t}^{(k)} \mathbf{t}_k + \mathbf{t}_{t_c \rightarrow t}^{(k)}, \\
        \mathbf{R}_k(t) &= \mathbf{R}_{t_c \rightarrow t}^{(k)} \mathbf{R}_k.
    \end{aligned}
    \label{eqn:rigid_transform}
\end{equation} 
Note that attributes like scaling $\mathbf{s}_k$, opacity $o_k$, and color $\mathbf{c}_k$ remain invariant over time to ensure temporal structural consistency. Furthermore, Different from our spatio-temporal alignment, MonoFusion performs a scale-alignment strategy using DUSt3R as the metric reference to achieve depth alignment with robust scale. Specifically, a global scale factor is estimated via the median ratio between DUSt3R and MoGE depth maps, computed exclusively over static background regions identified by SAM2.

\subsection{Evaluation Details}
As noted in the main text, the evaluation protocol for dynamic regions in MonoFusion~\cite{Wang2025ICCVa} is only vaguely described, and the official code lacks the corresponding evaluation scripts. To ensure a fair and reproducible comparison, we re-implement the evaluation pipeline for foreground dynamic objects. 

Specifically, we utilize SAM3~\cite{Carion2026ICLR} to generate precise foreground masks for each frame. We then compute the bounding box of the masked foreground region and extract the corresponding pixels. We found that simply calculating the PSNR of foreground pixels cannot reproduce the dynamic-region metrics reported in Table 2 of the MonoFusion paper. We thus pad the cropped foreground with a pure white background according to a specific scale ratio, and use these images for dynamic region evaluation. 

Since photometric metrics computed this way are highly sensitive to the proportion of background pixels, we conducted a sensitivity analysis on various padding ratios. Our investigation revealed that the choice of padding significantly shifts the quantitative results. Ultimately, we selected the padding ratio that yields metrics most consistent with the baseline values reported in the original MonoFusion paper.

\begin{table}[tbp]
\centering
\caption{Sensitivity analysis of the foreground padding ratio. We evaluate the impact of different padding scales on photometric metrics for dynamic objects. The ratio $1.4\times$ is selected as it yields results most consistent with the baseline values reported in MonoFusion~\cite{Wang2025ICCVa}.}
\label{tab:sensitivity_padding}
\begin{tabular}{lcccc}
\toprule
Padding Ratio & PSNR $\uparrow$ & SSIM $\uparrow$ & LPIPS $\downarrow$  \\
\midrule
$1.0\times$  &25.94 & 0.924 & 0.074  \\
$1.1\times$              & 27.20& 0.943 & 0.060 \\
$1.2\times$              &28.25 & 0.955 & 0.049  \\
$1.4\times$              & 29.48 & 0.958 & 0.040\\
\midrule
MonoFusion$\dagger$   &29.71 	& 0.947 &	0.017  \\
\bottomrule
\end{tabular}
\end{table}
\subsection{Losses}
In depth alignment stage, we set $\lambda_s =4$, $\lambda_m=5$, $\lambda_o=5$ and $\lambda_{ssi}=1$ and optimize for 10000 iterations in all experiments. When training different Gaussian splatting backends, we follow official training configurations of respective baselines. 

\section{Qualitative Analysis of Individual Components}
\begin{figure}[t]
    \centering
    \includegraphics[width=\linewidth]{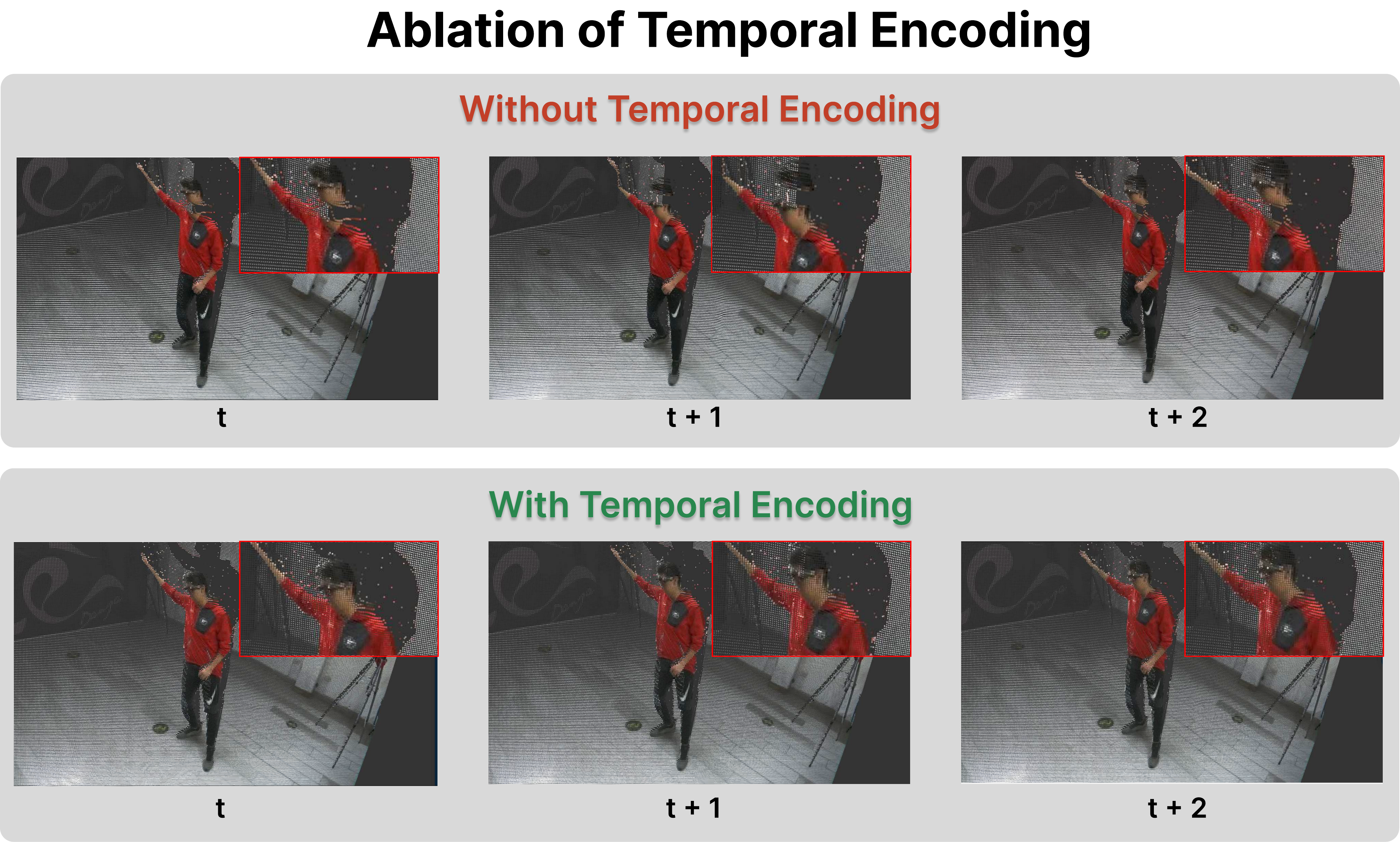}
    \caption{\textbf{Qualitative Analysis of temporal encoding}. We unproject depth maps from one view to demonstrate the effectiveness of our temporal encoding.}
    \label{fig:temporal-encoding-ablation}
\end{figure}

\begin{figure}[tbh]
    \centering
    \includegraphics[width=\linewidth]{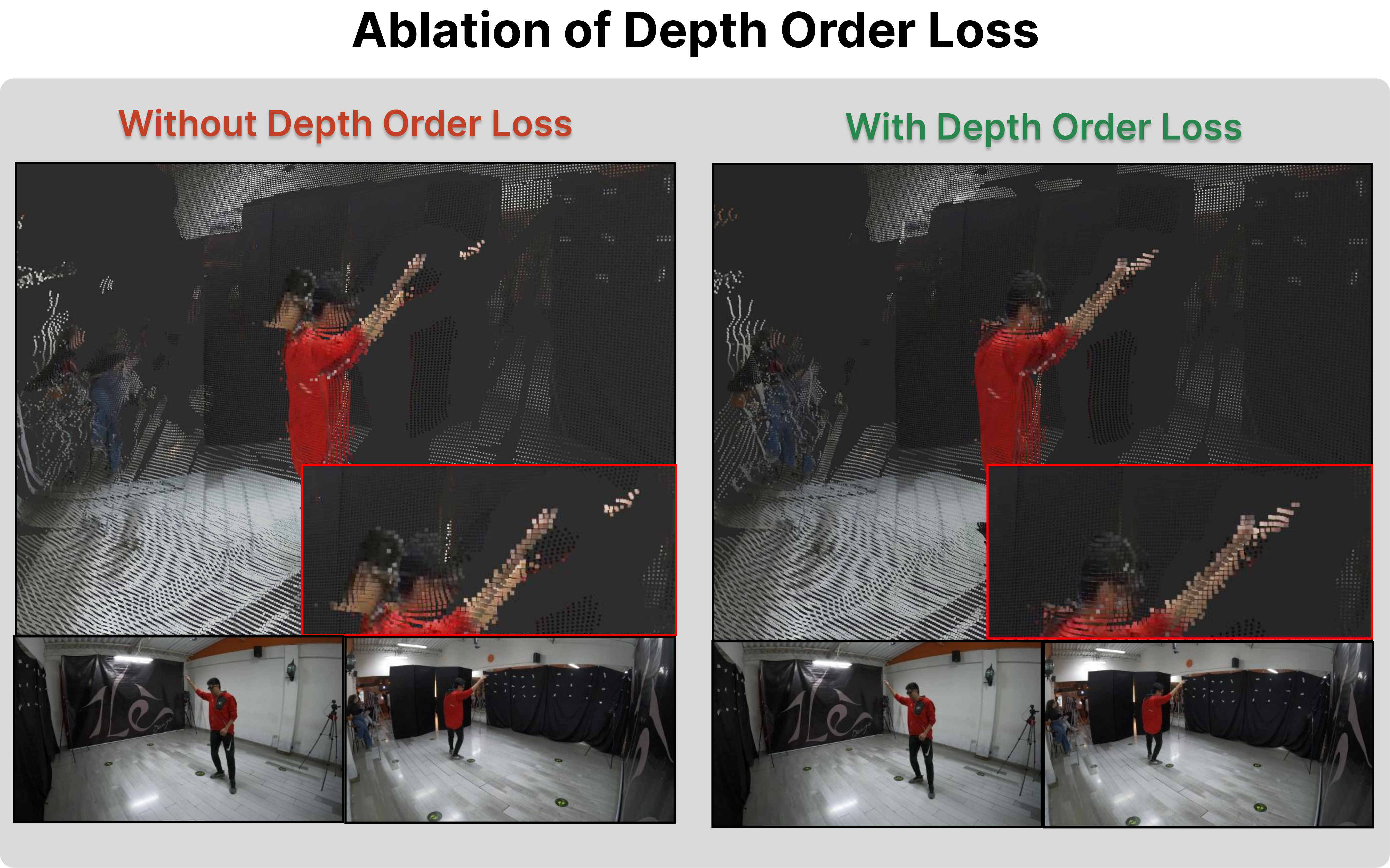}
    \caption{\textbf{Qualitative Analysis depth order loss}. For better understanding, we use depth unprojections from 2 opposed views to highlight multi-view consistency.}
    \label{fig:depth-order-ablation}
\end{figure}

To further validate the efficacy of our proposed framework, we provide a qualitative ablation analysis of two core components: Temporal Encoding and Depth Order Loss. Specifically, we validate depth quality qualitatively by unprojecting depth maps into 3D space. As shown in the Figures, temporal encoding consistently improves depth quality across time (\cref{fig:temporal-encoding-ablation}), while depth order loss significantly refines the multi-view geometry and removes floaters (\cref{fig:depth-order-ablation}).

\section{Direct Evaluation of Depth Alignment Quality}
\label{appx:direct_alignment_metrics}

To further validate that our method improves the geometric consistency of the aligned depth itself, we introduce three direct alignment-quality metrics that are independent of the downstream Gaussian-splatting backend.

First, we compute \textit{cross-view reprojection consistency}, which measures the mean $\ell_1$ error of pixel correspondences induced by the aligned depth across the four exocentric views. This metric directly evaluates whether the aligned depths from different views project to mutually consistent image locations. Second, we report the \textit{flow-warped depth error}, defined as the AbsRel depth error after warping adjacent frames with optical flow. This metric measures the temporal stability of the aligned depth across consecutive frames. Third, we evaluate the \textit{depth-order violation rate}, which measures the fraction of pixel pairs from two different views whose relative depth ordering becomes inconsistent after reprojection. This metric captures whether the aligned depth preserves cross-view ordinal geometry.

As shown in Table~\ref{tab:direct_alignment}, our method consistently improves all three direct alignment metrics on Ego-Exo4D. Compared with MAtCha and MonoFusion, our approach achieves lower cross-view reprojection error, lower flow-warped depth error, and a substantially lower depth-order violation rate. These results demonstrate that the proposed alignment strategy does not merely improve the final Gaussian-splatting reconstruction, but also produces geometrically more consistent depth before any rendering optimization.
\begin{table}[t]
  \centering\scriptsize\setlength{\tabcolsep}{4pt}
  \caption{
  Direct evaluation of depth alignment quality on Ego-Exo4D. 
  Our method improves cross-view reprojection consistency, temporal depth stability, and cross-view depth-order consistency without relying on Gaussian splatting optimization.
  }
  \label{tab:direct_alignment}
  \begin{tabular}{@{}l|ccc@{}}
  \toprule
  Method & Reproj. Consistency$\downarrow$ & Flow-warped Depth$\downarrow$ & Order Violation$\downarrow$ \\
  \midrule
  MAtCha     & 0.585 & 0.034 & 9.75\% \\
  MonoFusion & 0.648 & 0.016 & 28.10\% \\
  Ours       & \textbf{0.536} & \textbf{0.015} & \textbf{4.84\%} \\
  \bottomrule
  \end{tabular}
\end{table}

\section{Additional Scenes}
\label{appx:additional_scenes}

To further evaluate the generalization ability of our method across different dynamic scenes, we conduct an additional experiment on a ``hugging'' scene from the Harmony4D dataset~\cite{khirodkar2024harmony4d}. This scene contains close human-human interaction and therefore introduces challenging occlusions and non-rigid motion. We use 4 input cameras as was done for EgoExo4D and EgoHuman.

As shown in Table~\ref{tab:harmony4d}, our method improves the reconstruction quality significantly. This result suggests that the proposed alignment strategy generalizes beyond the EgoHuman sequences and can benefit reconstruction in different multi-human dynamic scenes.

\begin{table}[t]
\centering
\scriptsize
\setlength{\tabcolsep}{4pt}
\caption{
\textbf{Experiment on an additional Harmony4D scene.}
}
\label{tab:harmony4d}
\begin{tabular}{@{}l|ccc@{}}
\toprule
Method & PSNR$\uparrow$ & SSIM$\uparrow$ & LPIPS$\downarrow$ \\
\midrule
MonoFusion & 16.83 & 0.846 & 0.514 \\
MonoFusion + Ours     & \textbf{18.49} & \textbf{0.894} & \textbf{0.440} \\
Def-2DGS & 16.82 & 0.825 & 0.476 \\
Def-2DGS + Ours     & \textbf{19.15} & \textbf{0.906} & \textbf{0.331} \\
\bottomrule
\end{tabular}
\end{table}

\section{More Ablations}

\subsection{Ablation on Novel View Synthesis}
We ablate our design choices on novel view synthesis of the EgoHuman dataset. As shown in Table~\ref{appx:tab:ablation}, our proposed components improve novel view synthesis progressively.
\begin{table*}[t]
\scriptsize
  \centering
  \caption{\textbf{Progressive ablation of alignment components}
    on EgoHuman on two Gaussian splatting backends.}
  \label{appx:tab:ablation}
  \begin{tabular}{l|ccc|ccc}
    \toprule
    
    & \multicolumn{3}{c}{Def-GS} 
    & \multicolumn{3}{c}{MonoFusion} \\  
    Configuration
    & PSNR~$\uparrow$ & SSIM~$\uparrow$ & LPIPS~$\downarrow$
    & PSNR~$\uparrow$ & SSIM~$\uparrow$ & LPIPS~$\downarrow$ \\
    \midrule
    (a) Per-frame MAtCha                  & 20.93 & 0.830 & 0.305 & 19.54 & 0.566 & 0.516 \\
    (b) + residual weights (ours)         & 21.80 & 0.804 & 0.309 & 19.59 & 0.590 & 0.514 \\
    (c) + $\mathcal{L}_\text{order}$      & 22.15 & 0.815 & 0.306 & 19.94 & 0.583 & 0.506 \\
    (d) + $\mathcal{L}_\text{ssi}$ (full) & 22.76 & 0.856 & 0.277 & 20.41 & 0.612 & 0.492 \\
    \bottomrule
  \end{tabular}
\end{table*}

\subsection{Ablations on Residual Rank and $\mathcal{L}_\text{order}$}

We further study the effect of the residual rank $R$ and the formulation of the depth order regularization $\mathcal{L}_\text{order}$ on Ego-Exo4D with Def-2DGS + our full model.

\paragraph{Residual rank $R$.}
We evaluate $R \in \{4, 8, 16\}$ to control the capacity of the residual motion representation. As shown in Table~\ref{tab:ablation_rank}, increasing $R$ improves performance up to $R=8$, after which the performance saturates and slightly degrades, indicating a trade-off between modeling capacity and overfitting.

\begin{table}[t]
\centering
\scriptsize
\setlength{\tabcolsep}{5pt}
\caption{Ablation on residual rank $R$ on Ego-Exo4D (Def-2DGS + Ours).}
\label{tab:ablation_rank}
\begin{tabular}{c|ccc}
\toprule
$R$ & PSNR$\uparrow$ & SSIM$\uparrow$ & LPIPS$\downarrow$ \\
\midrule
4  & 33.76 & 0.969 & 0.047 \\
8  & 34.66 & 0.963 & 0.049 \\
16 & 34.51 & 0.966 & 0.045 \\
\bottomrule
\end{tabular}
\end{table}

\paragraph{$\mathcal{L}_\text{order}$.}
We compare the proposed hinge formulation with a symmetric $L_1$ variant to study the robustness of the depth-order constraint. As shown in Table~\ref{tab:ablation_order}, both formulations achieve comparable results, with the hinge loss providing slightly better PSNR, suggesting stability across penalty designs.

\begin{table}[t]
\centering
\scriptsize
\setlength{\tabcolsep}{5pt}
\caption{Ablation on $\mathcal{L}_\text{order}$ formulation on Ego-Exo4D (Def-2DGS + Ours).}
\label{tab:ablation_order}
\begin{tabular}{c|ccc}
\toprule
Loss type & PSNR$\uparrow$ & SSIM$\uparrow$ & LPIPS$\downarrow$ \\
\midrule
Hinge (ours) & \textbf{34.66} & 0.963 & 0.049 \\
Symmetric $L_1$ & 34.44 & \textbf{0.966} & \textbf{0.047} \\
\bottomrule
\end{tabular}
\end{table}

\subsection{Ablation on Different Camera Baselines}
\label{appx:ablation_cam_baselines}

We further evaluate the robustness of our method under different sparse-view camera baselines. Specifically, we conduct additional experiments on EgoHuman with 3 and 6 input cameras, which roughly correspond to $120^\circ$ and $60^\circ$ camera spacing, respectively. For evaluation, we use the cameras located between the selected input views as held-out novel views, and adopt Def-2DGS as the reconstruction backend.

As shown in Table~\ref{tab:camera-baseline}, our method consistently improves the reconstruction quality under both settings. These results demonstrate that our method remains effective under different camera baselines, including the more challenging 3-camera setting with wider view spacing.

\begin{table*}[tb]
  \scriptsize
  \centering
  \caption{Evaluation under different camera baselines on EgoHuman using Def-2DGS as the backend. The 3- and 6-camera settings roughly correspond to $120^\circ$ and $60^\circ$ camera spacing, respectively.}
  \label{tab:camera-baseline}
  \begin{tabular}{c|l|ccc}
    \toprule
    \#Cams & Method 
    & PSNR$\uparrow$ & SSIM$\uparrow$ & LPIPS$\downarrow$ \\
    \midrule
    \multirow{2}{*}{3} 
    & Def-2DGS & 15.14 & 0.485 & 0.630 \\
    & Def-2DGS + Ours & \textbf{18.91} & \textbf{0.731} & \textbf{0.417} \\
    \midrule
    \multirow{2}{*}{6} 
    & Def-2DGS & 18.42 & 0.645 & 0.468 \\
    & Def-2DGS + Ours & \textbf{21.79} & \textbf{0.842} & \textbf{0.271} \\
    \bottomrule
  \end{tabular}
\end{table*}

\begin{figure}[tb]
    \centering
    \includegraphics[width=0.9\linewidth]{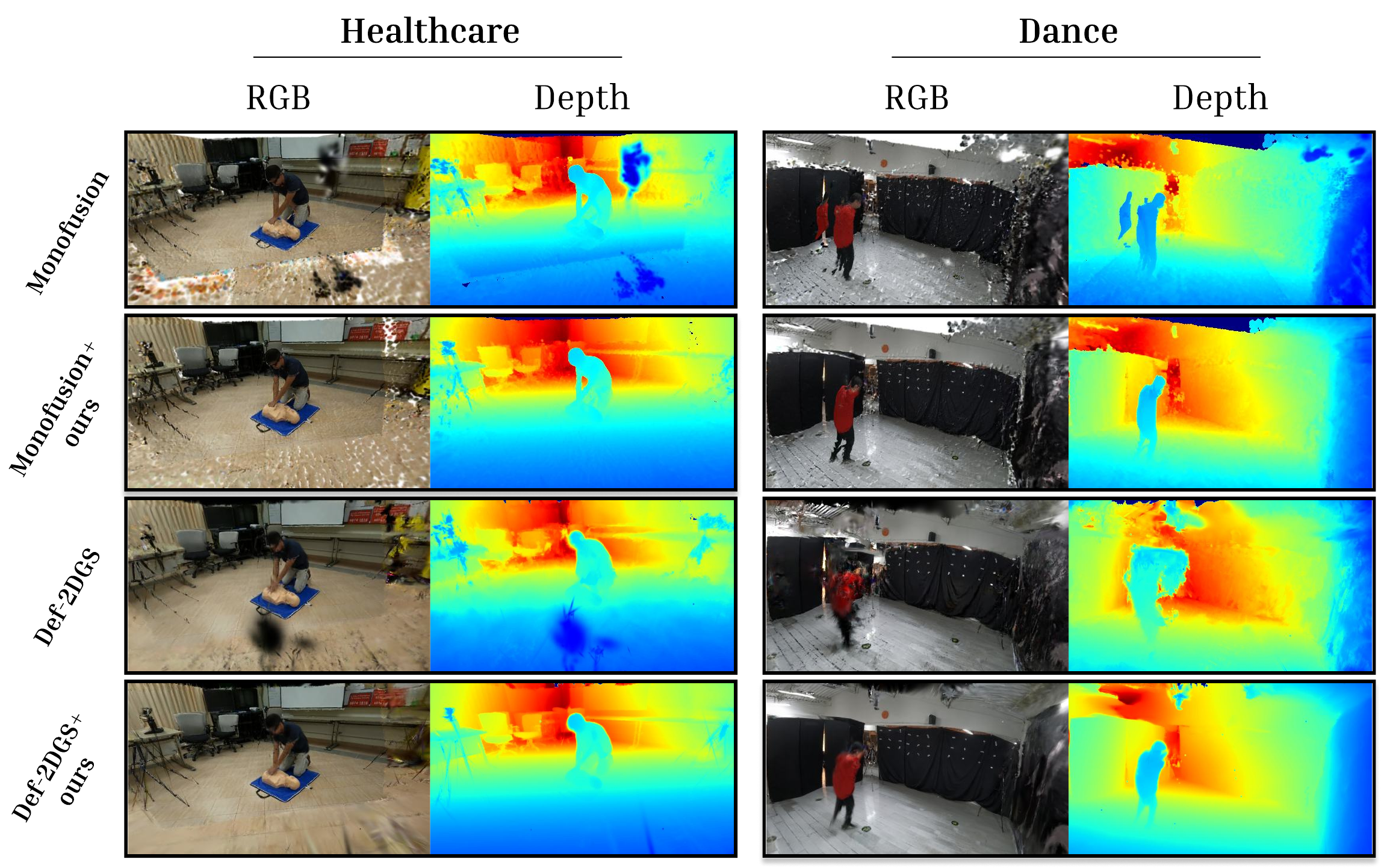}
    \caption{\textbf{More qualitative results of novel view synthesis on EgoExo4D.} }
    \label{fig:more-qualitative}
\end{figure}

\section{More Qualitative Results}
To further demonstrate the robustness of our method, we provide additional qualitative results on novel views in Fig~\ref{fig:more-qualitative}. Results better demonstrate that our method robustly improves the geometry quality of reconstructed scenes.


\begin{figure}[h]
    \centering
    \includegraphics[width=0.8\linewidth]{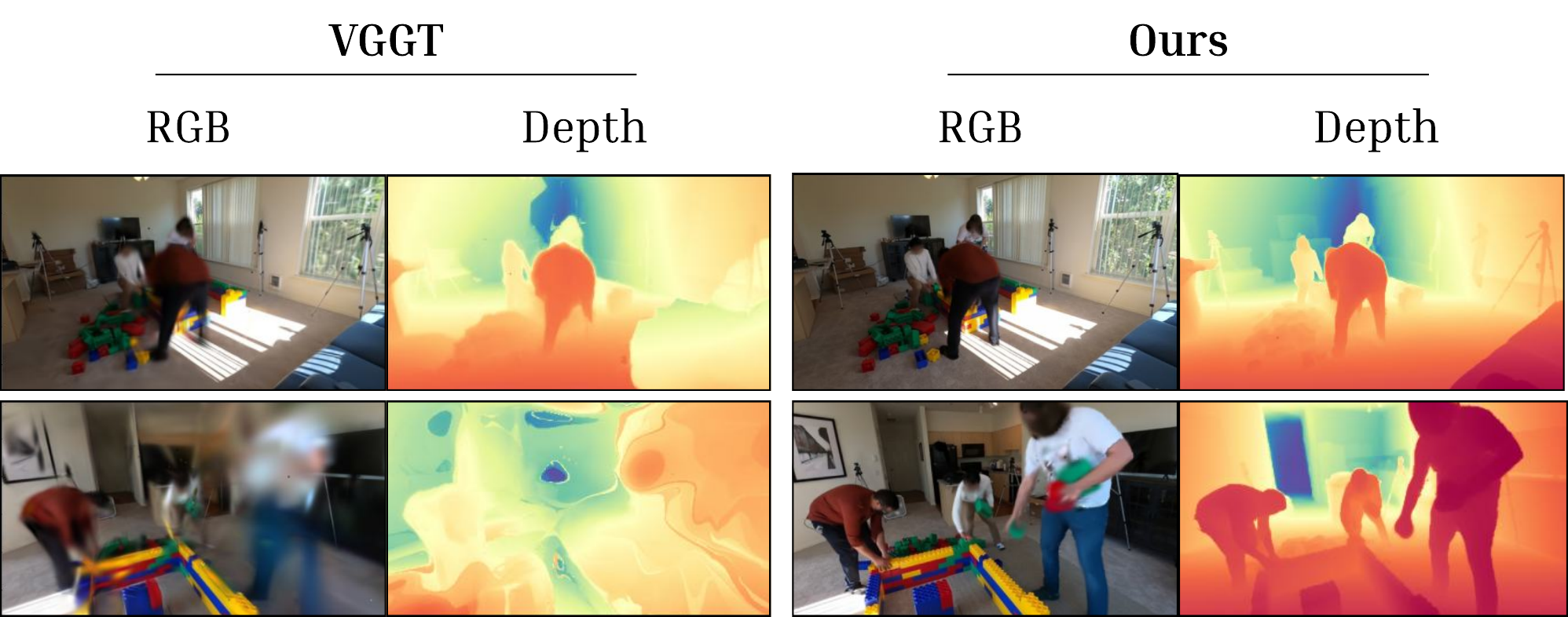}
    \caption{\textbf{We initialize and supervise Deformable 2DGS on EgoHuman using depth priors predicted by VGGT.}}
    \label{fig:VGGT-qualitative}
\end{figure}
\section{CMU Panoptic Dataset}
We tried to run MonoFusion on the CMU Panoptic dataset~\cite{Joo_2017_TPAMI} to reproduce the reported metrics in the original MonoFusion paper; however, the first preprocessing step (DUSt3R~\cite{Wang2024CVPR}) always fails under the condition of 4 input views, even with known camera parameters. We hypothesize that this could be due to the pure white background of the capture dome. We also notice that similar issues were reported on the official GitHub repository, with no real solution at the time of submission.

\section{VGGT Priors}
We tried to initialize and supervise Deformable 2DGS using the VGGT model to evaluate the quality of VGGT depth priors. As shown in Fig~\ref{fig:VGGT-qualitative}, the depth priors of VGGT are suboptimal for robust scene reconstruction, leading to vague rendering results.


%
%
\bibliographystyle{splncs04}
\bibliography{main}
\end{document}